\documentclass[preprint,12pt,num]{elsarticle}
\usepackage{algorithmic}
\usepackage{amssymb}
\usepackage{amsthm}
\usepackage{mathtools}
\usepackage{easyReview}
\usepackage{lineno}
\usepackage{float}
\usepackage{amsmath}
\usepackage{physics}
\usepackage{lscape}
\usepackage{graphicx}
\usepackage{booktabs}
\usepackage{threeparttable}
\usepackage{footnote}
\usepackage{geometry}
\usepackage{longtable}

\usepackage[ruled,vlined]{algorithm2e} 
\usepackage[normalem]{ulem}

\begin{document}

\begin{frontmatter}

	\title{Jump Diffusion-Informed Neural Networks with Transfer Learning for Accurate American Option Pricing under Data Scarcity}

	%\title{Stochastic Jump Diffusion Process Informed Neural Network with Partial Integro-differential Equation Enhanced Transfer Learning for American Option Pricing under Scarce Data Condition}

	\author[label1]{Qiguo Sun \corref{cor1}}
	\ead{sunqiguo1992@gmail.com}
	\author[label2]{Hanyue Huang}
	\ead{huanghanyue01@gmail.com}
	\author[label1]{Xibei Yang}
	\ead{jsjxy\_yxb@just.edu.cn}
	\author[label1]{Yuwei Zhang}
	\ead{zhangyuwei@seu.edu.cn}

	\affiliation[label1]{organization={School of Computer Science, Jiangsu University of Science and Technology},
		%%             addressline={},
		city={Zhenjiang},
		postcode={212003},
		state={Jiangsu Province},
		country={China}}
	\affiliation[label2]{organization={Technical University of Munich},
		%%             addressline={},
		city={Munich},
		postcode={80333},
		state={Bavaria},
		country={Germany}}

	\cortext[cor1]{Corresponding author}

	\begin{abstract}
		Option pricing models, essential in financial mathematics and risk management, have been extensively studied and recently advanced by AI methodologies. However, American option pricing remains challenging due to the complexity of determining optimal exercise times and modeling non-linear payoffs resulting from stochastic paths. 
		Moreover, the prevalent use of the Black-Scholes formula in hybrid models fails to accurately capture the discontinuity in the price process, limiting model performance, especially under scarce data conditions.
		To address these issues, this study presents a comprehensive framework for American option pricing consisting of six interrelated modules, which combine nonlinear optimization algorithms, analytical and numerical models, and neural networks to improve pricing performance.
		Additionally, to handle the scarce data challenge, this framework integrates the transfer learning through numerical data augmentation and a physically constrained, jump diffusion process-informed neural network to capture the leptokurtosis of the log return distribution.  
		To increase training efficiency, a warm-up period using Bayesian optimization is designed to provide optimal data loss and physical loss coefficients.		
		Experimental results of six case studies demonstrate the accuracy, convergence, physical effectiveness, and generalization of the framework. Moreover, the proposed model shows superior performance in pricing deep out-of-the-money options.
	\end{abstract}


\end{frontmatter}

%% \linenumbers

%% main text
\section{Introduction}

Options are fundamental financial derivatives widely employed for risk management.
The movement of option prices follows a stochastic process influenced by various factors such as the price process of the underlying assets $(S_t)$, the strike price $(K)$, the time-to-maturity $(T)$, the option type (American or European; Put $(P)$ or Call $(C)$ options), and numerous macroeconomic and market factors.
Therefore, option pricing is a challenging task that has been a focus of financial mathematics for decades since the proposal of the famous Black-Scholes (BS) formula \cite{black1973pricing}.

\begin{figure}[H]
	\centering
	\includegraphics[trim=0cm 0cm 0cm 0cm,clip=true,width=10cm]{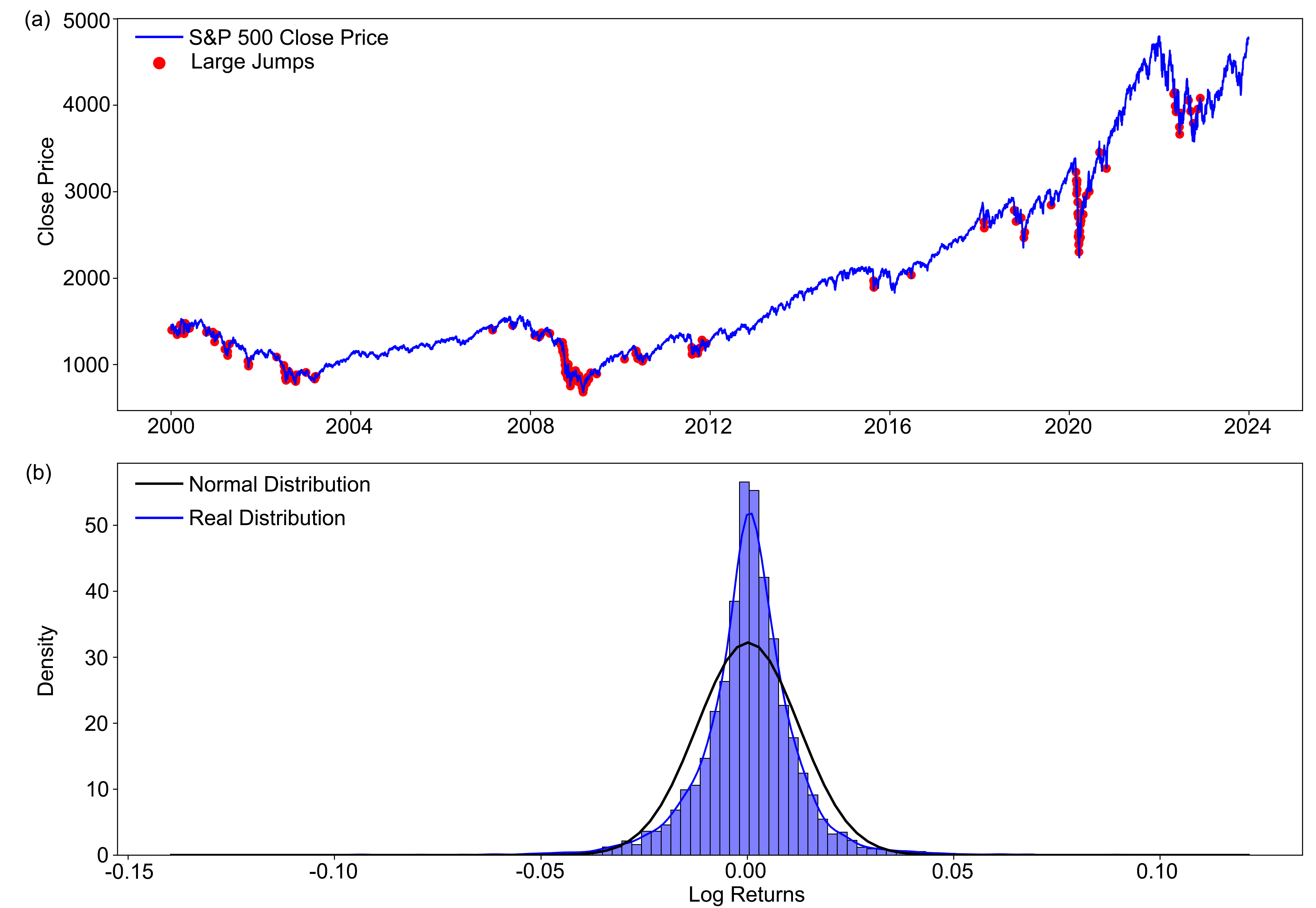}
	\vspace*{-3mm}
	\caption{(a) Close price of S\&P 500 index with large jumps ( $|log(dS)|> 0.03$) highlighted in red; and (b) distribution of log returns and fitted Normal in black. }
	\label{Fig:abstract}
\end{figure}

The BS model models stochastic paths of underlying assets as Geometric Brownian Motion (GBM) by assuming constant volatility and interest rates, as well as log-normal price distributions.
BS model is widely employed in research and practical applications as it can provide analytical solutions for European call and put options pricing.
However, the continuous nature of GBM has difficulty in capturing real market data, which generally exhibits jumps and spikes in price process (See Figure \ref{Fig:abstract}(a)) and higher peak and fatter tails in log returns (See Figure \ref{Fig:abstract}(b)).

To address these limitations, Merton \cite{merton1976option} developed the first jump diffusion model by adding a composite Poisson process to the GBM to model the discontinuous price process.
Heston further developed a stochastic volatility model \citep{heston1993closed} with two Stochastic Differential Equations (SDEs), where the second SDE describes the volatility process assumed to follow an Ornstein-Uhlenbeck process.
Additionally, Lévy Processes \citep{ken1999levy} and the Variance Gamma model \citep{madan1990variance, madan1998variance} were developed to capture the skewness and kurtosis of return series, thus providing a more flexible and accurate representation of price dynamics.

Recently, machine learning and AI techniques offer significant potential in constructing option pricing models due to their strong expressiveness and computational efficiency, though they face challenges in physical effectiveness and generalization across market conditions.
In \cite{IVASCU2021113799}, the performance of various prevailing data-driven approaches applied to option pricing tasks was investigated.
The results show that machine learning approaches outperform classical parametric models. Moreover, data-driven methods integrating traditional parametric approaches have been an active research domain.
In \cite{jang2021deepoption}, an option pricing and delta-hedging framework called DeepOption is proposed, which utilizes data fusion methods to integrate estimations from various parametric option pricing models.
Experimental results on real datasets of S\&P500, EuroStoxx50, and Hang Seng Index demonstrate the effectiveness of DeepOption compared to parametric methods and other machine learning methods.
\cite{cao2023gamma} developed an option hedging model based on a reinforcement learning framework, in which the price and volatility are modeled using two SDEs.
An option pricing model integrating the parametric BS formula into an equilibrium model was developed in \cite{sheybani2021equilibrium}.
In contrast to previous works using implied volatility derived from historical price data, this new approach computes equilibrium volatility derived from day-ahead electric use future spot prices.
In addition, the Physics-Informed Neural Networks (PINNs) framework has been employed to construct option pricing models in \cite{dhiman2023physics}.

Despite these data-driven models can effectively overcome the computational inefficiencies of numerical models, several unsolved challenges remain.
First, for American option pricing tasks, data-driven methods struggle to maintain accuracy on unseen datasets due to the difficulty in determining optimal exercise times and modeling non-linear payoffs resulting from stochastic paths.
Second, the prevalent use of the BS formula in hybrid models fails to accurately capture the leptokurtic nature of return distributions.
Without hard constraints on parameters, these models can produce extreme and unrealistic values for the GBM process.
Third, option data is expensive and less accessible compared to other financial data \citep{jang2021deepoption, chan2017machine}, while few studies focus on developing data-efficient algorithms capable of handling very scarce data conditions.

To address these challenges, this article proposes a comprehensive American option pricing framework consisting of six interrelated modules:
(1) Parameter Estimation Module: tune parameters based on market data for parameter assignment in analytical and numerical models; define constraints to control parameter update  in the pricing module;
(2) American Option Pricing Module: decompose American option price into European price and early exercise premiums and use neural networks to parametrize them and provide estimation;
(3) Analytical Merton Model: define Merton jump diffusion process-based European option pricing model to support physical loss calculations;
(4) Loss Coefficients Optimization Module: employ Bayesian Optimization to enhance training efficiency;
(5) Numerical PIDE Solver Module: generate synthetic datasets based on a partial integro-differential equation solver for data augmentation;
and (6) Training Module: integrate transfer learning, design the total loss function, and employ various techniques to improve training performance.

The main contributions of this research are threefold:
\begin{itemize}
	\item \textbf{Comprehensive Framework for American Option Pricing}
	      The framework integrates six interrelated modules, combining nonlinear optimization algorithms, the analytical model, the numerical differential equation solver, neural networks, and transfer learning to achieve parameter tuning, physical constraints, and data augmentation,
	      thereby enhancing pricing performance and physical effectiveness.
	\item \textbf{Efficient Handling of Scarce Data Conditions}
	The framework integrates transfer learning based on numerical data augmentation and a jump diffusion process-informed neural network to effectively capture the leptokurtosis of return series. 
	This physics- and data-enhanced framework significantly improves model performance and generalization under scarce data conditions. 
	\item \textbf{Training Efficiency}
	      Designing a warm-up period using Bayesian optimization to provide loss coefficients to balance data loss and physical loss to enhance training efficiency.
\end{itemize}

The remainder of this paper is organized as follows:
Section 2 reviews the relevant literature; Section 3 presents the methodology; Section 4 details the implementation;
Section 5 presents the experimental results and analysis; Section 6 concludes this work;
Finally, the derivation of the formula for the analytical solution and the matrix design of the numerical solution are shown in the Appendix.

\section{Literature Review}
\subsection{Option Pricing Under Parametric Models}

Research on parametric models for option pricing has made significant progress over the past half-century.
The seminal work by Black and Scholes \cite{black1973pricing} introduced the BS formula, providing an analytical solution for European options pricing.
The BS model assumes that asset prices are log-normally distributed and that both volatility and interest rates are constant.
Despite the tractability of the BS model in real applications, it does not account for common phenomena such as jumps and fat tails in return series.

To address these issues, Merton \cite{merton1976option} included jumps term defined as composite Poisson process to the GBM, resulting in the Merton jump diffusion model.
This model captures sudden, large changes in prices, thus capturing the leptokurtic nature of asset return distributions, providing a more accurate framework for modeling stochastic paths.
Moreover, stochastic volatility models were developed in \cite{heston1993closed} to simultaneously model SDEs of prices and volatility.
The Heston model also provides an analytical solution for European option pricing based on Fourier transform techniques.
Concurrently, the Variance Gamma (VG) models, a pure jump process with finite moments, were developed in \cite{madan1990variance, madan1998variance}.
VG models excel in capturing skewness and kurtosis in return series and are widely used in option pricing and hedging.

A more flexible jump diffusion model was developed in \cite{cai2011option}, where the jump size has a mixed-exponential distribution.
Moreover, the analytical pricing formula for path-dependent options is derived by explicitly solving a high-order integro-differential equation.
After parameter calibration, the model demonstrated effectiveness in pricing SPY options.
In \cite{cheang2012modern}, Radon-Nikodým derivative process was used to change the market measure to an equivalent martingale measure, and thus developing a partial integro-differential equation (PIDE) for option pricing.
The effectiveness of the Merton jump diffusion model and the BS model in capturing the leptokurtic feature of log-returns and the \textit{volatility smile} is investigated in \cite{gugole2016merton}.
The results show that the Merton jump diffusion model outperforms GBM approaches.
\cite{zhou2020pricing} employed the Merton jump diffusion model to construct a valuation framework for pricing equity warrants, with experimental results proving their model outperforms the BS model.

Despite the active development of the jump diffusion models, there is few hybrid model combining parametric jump diffusion model with neural networks for American option pricing.

\subsection{Option Pricing Based on Machine Learning}

Machine learning and AI techniques offer significant potential in constructing option pricing models. They provide stronger expressiveness than analytical models and are more computationally efficient compared to numerical models.
However, their limitations include challenges in ensuring physical effectiveness and generalizing across different market conditions.
A comprehensive study was conducted in \cite{IVASCU2021113799} to investigate the performance of various option pricing models based on machine learning algorithms, including neural networks, support vector regressions, genetic algorithms, XGBoost, and LightGBM.
The dataset comprises European call options with WTI crude oil futures as the underlying assets. They found that machine learning methods generally outperform parametric approaches, with XGBoost outperforming other machine learning models.
A gated neural networks-based approach was proposed in \cite{yang2017gated} to price European options, which demonstrated good generalization ability.

Furthermore, integrating parametric models into data-driven methods is becoming a new trend in option pricing model development.
In \cite{das2017new}, a hybrid model was proposed which combines the BS model with machine learning models. Experimental results showed the hybrid model improved mean absolute error (MSE) by more than 76\% across four datasets.
The DeepOption model was proposed in \cite{jang2021deepoption}. DeepOption adopts a data fusion framework that processes distilled data from various parametric models, achieving a very low RMSE of 11.0 on the dataset of S\&P 500 call options.
In \cite{cao2023gamma}, an option hedging model based on reinforcement learning is developed. The model simulates asset price and volatility movements as two Wiener processes with constant correlation.
Additionally, an analytical module was designed to estimate implied volatility.
For option pricing in electricity markets, \cite{sheybani2021equilibrium} innovatively combined the BS model with an equilibrium model to construct the option pricing model, calculating option prices using the BS formula with equilibrium volatility.
Moreover, in \cite{dhiman2023physics}, Dhiman et al. first proposed an option pricing model based on the PINN framework, integrating the BS formula to guide training, highlighting that parameter optimization plays a key role in model performance and robustness.

Previous research provides significant motivation in integrating data-driven models with traditional parametric approaches to simultaneously improving model accuracy and physical effectiveness.
However, most research integrate BS formula into the hybrid models, which fails to accurately capture the leptokurtic nature of return distributions in real data.

\section{Methodology}

This subsection briefly introduces the mathematical design of the proposed framework, with detailed derivations of the analytical solution for the Merton pricing model and the matrix design of the numerical solver provided in the Appendix.
American option pricing is an optimal stopping problem, which aims to find the optimal exercise time $\tau$ that maximizes the expectation:
\begin{equation}
	V(t,s) =  \sup_{\tau \in [t,T]} \mathbb{E}^{\mathbb{Q}}[ e^{-r(\tau-t)} \Phi(S_{\tau}) | S_t=s ],
	\label{eq:American_ops}
\end{equation}
where $\Phi(\cdot)$ is the payoff function. Since $\tau$ is a random variable dependent on the price paths, at each time step of a given path, the pricing algorithm needs to determine if it is optimal to exercise.
Therefore, under GBM process, given initial condition $V(T,s)=\Phi(s)$, the option pricing problem involves solving the following partial differential equation:
\begin{equation}
	\max \left\{  \frac{\partial V(t,s)}{\partial t}  + rs \frac{\partial V(t,s)}{\partial s} + \frac{1}{2} \sigma^2 s^2 \frac{\partial^2  V(t,s)}{\partial s^2} - rV(t,s), \Phi(S_{s}) - V(t,s)  \right\}.
	\label{eq:American_ops2}
\end{equation}

It can be seen from Equation (\ref{eq:American_ops2}) that the pricing algorithm defines a free boundary that separates the stopping region and the continuation region.
In this work, we consider an alternative approach: the fair price of American options, $V(T,S)$, is assumed to be a linear combination of the European option price, $V_E(T, S)$, and a premium for early exercise, $\Pi(T, S)$.
Both components are parameterized using neural networks, with the coefficients of the two terms denoted by  $\eta$ and $\gamma$.
Therefore, the American option price can be calculated by:
\begin{equation}
	V(T, S) = \eta \cdot V_E(T, S) + \gamma \cdot \Pi(T, S)
	\label{eq:real_weights}
\end{equation}

Then, the model training can be defined as optimizing the following formula:
\begin{equation}
	\underset{\psi, \eta, \gamma}{\mathrm{argmin}} \quad \mathbb{E} \left( \left( \eta \cdot \hat{V}_E + \gamma \cdot \hat{\Pi} - V \right)^2 \right)
	\label{eq:expect_corrected}
\end{equation}
where $\psi$ indicate the parameters of the hybrid pricing model based on analytical solution and neural networks.
To enhance model flexibility and expressiveness, we use neural network parameterized models, $\tilde{V}_E$, to replace the analytical model, $\hat{V}_E$, for European option pricing, resulting in the following function:
\begin{equation}
	\underset{\theta, a}{\mathrm{argmin}} \quad \mathbb{E} \left( \left( a \cdot \tilde{V}_E + (1-a) \cdot \hat{\Pi} - V \right)^2 + \left( \tilde{V}_E  - V_E \right)^2 \right),
	\label{eq:expect2_corrected}
\end{equation}
where $\theta$ indicates the parameters of neural network parameterized American option pricing model, $a$ is a trainable parameter in the model with $a = \eta , 1-a= \gamma $.

Though observation of Equation (\ref{eq:expect2_corrected}), it can be found that the fisrt term calculates a data loss ($L_d$), while the second term calcultes a physics loss ($L_p$).
To improve training efficiency, we introduce coefficients to adjust the importance of $L_d$ and $L_p$ during training, so that the optimization task becomes:
\begin{equation}
	\underset{\theta, a,\alpha, \beta}{\mathrm{argmin}} \quad \mathbb{E} \left( \beta \left( a \cdot \tilde{V}_E + (1-a) \cdot \hat{\Pi} - V \right)^2 + \alpha \left( \tilde{V}_E  - V_E \right)^2 \right).
	\label{eq:expect3_corrected}
\end{equation}

We consider optimizing $\theta$ in Equation (\ref{eq:expect3_corrected}) conditioned on the hyperparameters $ a,\alpha, \beta$.
However, using grid search methods would lead to a computational burden.
Therefore, we design a warm-up period for coefficients optimization based on Bayesian optimization, which is presented later.
Finally, given the optimal set of hyperparameters, the total loss $L$ can be defined as:
\begin{equation}
	L = \frac{1}{N}\sum_{i=1}^{N} (\alpha \cdot L_p + \beta \cdot L_d),
	\label{eq:loss}
\end{equation}
where $N$ is the number of samples.

\section{System Framework}

\begin{figure}[H]
	\centering
	\includegraphics[trim=0cm 0cm 0cm 0cm,clip=true,width=15cm]{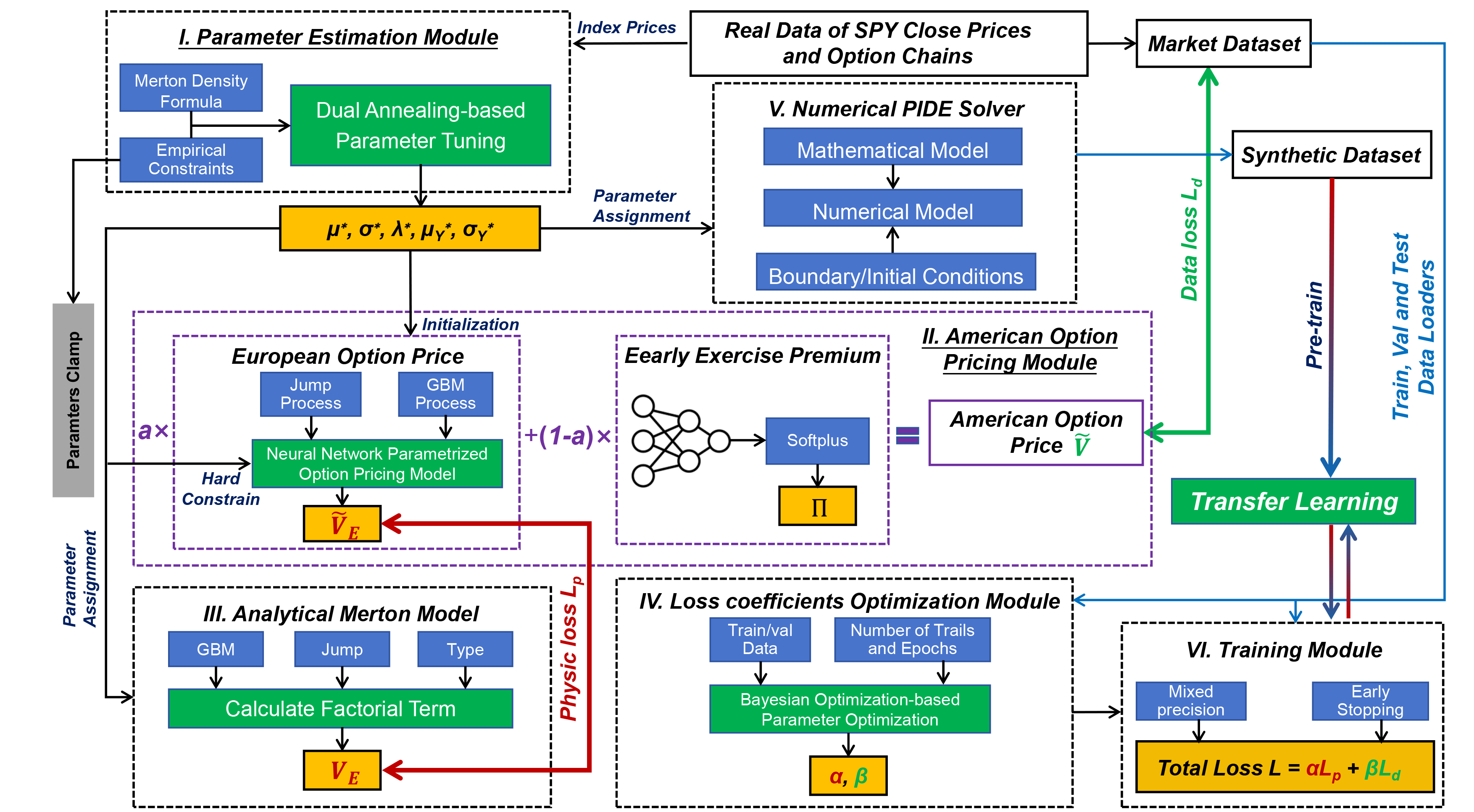}
	\vspace*{-3mm}
	\caption{Framework of PINN-Merton model}
	\label{Fig:framework}
\end{figure}

The propsed system framework is shown in Figure \ref{Fig:framework}, which contains six key modules. Their main functions can be summarized as follows:
\begin{itemize}
	\item \textit{I. Parameter Estimation Module}: Use real data to estimate optimal parameters based on the Dual Annealing algorithm and define parameter bounds to clamp parameter in Module \textit{II} within physics constraints.
	      The optimized parameters are used to initializing networks in Module \textit{II} and assign parameters in Modules \textit{III} and  \textit{V}.
	\item \textit{II. American Option Pricing Module}: Implementation of the American option price estimation by Equation (\ref{eq:american option price}).
	      Provide estimation of $\tilde{V}_E$ and $\tilde{V}$ for calculating  $L_p$ and $L_d$.
	\item \textit{III. Analytical Merton Model}: Implementation of the Merton jump diffusion model for European option pricing using parameters given by Module \textit{I}, providing $V_E$ to calculate $L_p$.
	\item \textit{IV. Loss Coefficients Optimization Module}: An efficient global optimization module based on the Bayesian Optimization to estimate optimal hyperparameters $\alpha, \beta$ for Module \textit{VI}.
	\item \textit{V. Numerical PIDE Solver Module}: Provide data augmentation based a numerical American option PIDE solver,
	      providing synthetic datasets for transfer learning in Module \textit{VI}.
	\item \textit{VI. Training Module}: Design main framework to train PINN-Merton and PINN-Merton-Transfer models, integrating various techniques to enhance training performance.
\end{itemize}

\subsection{I.Parameter Estimation Module}

To find global optima in complex landscapes, a parameter optimization module is designed in Algorithm \ref{algorithm parameters}.
This algorithm employs Dual Annealing optimization, a stochastic global optimization algorithm, to conduct the nonlinear optimization tasks.
Specifically, the objective function is defined based on the Merton density function ($f_{DM}$) given in \citep{tankov2003financial},
\begin{equation}
	f_{DM}( \mu, \sigma, \lambda, \mu_Y, \sigma_Y | (\mathbf{x}, T)) = \exp(-\lambda T) \sum_{k=0}^{K} \frac{(\lambda T)^k}{k!} \frac{\exp\left(-\frac{(\mathbf{x}- \mu T - k \mu_Y)^2}{2 (T \sigma^2 + k \sigma_Y^2)}\right)}{\sqrt{2 \pi (T \sigma^2 + k \sigma_Y^2)}}.
	\label{eq: Merton density}
\end{equation}
We modify this $f_{DM}$ to log likelyhood form as objective function $f_{LLDM}$ as,
\begin{equation}
	f_{LLDM}(\mu, \sigma, \lambda, \mu_Y, \sigma_Y  | (\mathbf{x}, T)) = -\sum_{i=1}^{N} \log(f_{DM}(x_i, T, \mu, \sigma, \lambda, \mu_Y, \sigma_Y)),
	\label{eq:log_likely_merton}
\end{equation}
where the parameter bounds $\mathbf{b}$ refer to \citet{matsuda2004introduction}.

\begin{algorithm}[H]
	\SetAlgoLined
	\KwIn{Log returns $\mathbf{x}$, Time period $T$, Bounds $\mathbf{B}$, Max iterations $K$, Tolerance $\epsilon$}
	\KwOut{Optimal parameters $\mu^*$, $\sigma^*$, $\lambda^*$, $\mu_Y^*$, $\sigma_Y^*$}
	\textbf{Initialize} Parameters $\mu, \sigma, \lambda, \mu_Y, \sigma_Y$ within bounds $\mathbf{b}$\\
	\textbf{Set} bounds $\mathbf{b}$ for parameters\\
	\textbf{Optimization using Dual Annealing:}\\
	\Indp
	\textbf{Step 1:} Define objective function $f_{LLDM}$.\\
	\textbf{Step 2:} Specify $\mathbf{b}$ for each parameter:\\
	$[(\mu_{\min}, \mu_{\max}), (\sigma_{\min}, \sigma_{\max}), (\lambda_{\min}, \lambda_{\max}), (\mu_{Y,\min}, \mu_{Y,\max}), (\sigma_{Y,\min}, \sigma_{Y,\max})]$ \\
	\textbf{Step 3:} Call Dual Annealing optimizer $f_{DA}$  in \textit{Scipy} with arguments: \\
	$\mu^*, \sigma^*, \lambda^*, \mu_Y^*, \sigma_Y^*$ $\leftarrow$ $f_{DA}(f_{LLDM}, \mathbf{b}, \text{args} = (\mathbf{x}, T), \text{maxiter} = K)$ \\
	\Indm
	\Return $\mu^*, \sigma^*, \lambda^*, \mu_Y^*, \sigma_Y^*$
	\caption{Parameter Optimization Based on Dual Annealing}
	\label{algorithm parameters}
\end{algorithm}

\subsection{II.American Option Pricing Module}

\textbf{European Option Price} \\
A neural network parameterized European Option pricing model for Merton jump diffusion process is designed to estimate $\tilde{V}_E$.
The jump term is composed of stochastic arrivals defined by Poisson process and jump amplitudes following a Gaussian distribution, defined as,\\
\begin{equation}
	\begin{cases}
		P[N_0=0] = 1,                                              \\
		P[N(t+\Delta t)- N(t)=1] = \lambda \Delta t + o(\Delta t), \\
		P[N(t+\Delta t)- N(t)>1] = o(\Delta t),                    \\
	\end{cases}
	\label{eq:Nt}
\end{equation}
where the intensity $\lambda$ is a trainable parameter. The design $P[N(t+\Delta t)- N(t)>1] = o(\Delta t)$ implies that the probability of multiple jumps within a small interval is negligible.
Moreover, the jump size ($Y_t$) follows a log-normal distribution and it is independent of $N_t$, defined as,\\
\begin{equation}
	\ln Y_t \sim \mathcal{N}(\mu_Y,\sigma_Y^2),
	\label{eq:yt}
\end{equation}
where $\mu_Y$ and $\sigma_Y^2$ are defined as trainable parameters.

Under the risk-neutral measure $\mathbb{Q}$ the European option pricing formula can be analytically expressed as,\\
For call options:
\begin{equation}
	\begin{aligned}
		\tilde{V}_{E} = \sum_{\nu = 0}^{\infty} e^{-\lambda T} \frac{(\lambda T)^{\nu}}{\nu!} \left( S e^{{\nu}\mu_Y} N(d_1^{\nu}) - K e^{-rT} N(d_2^{\nu}) \right) \\
		\approx  \sum_{0}^{n} e^{-\lambda T} \frac{(\lambda T)^n}{n!} \left( S e^{n\mu_Y} N(d_1^n) - K e^{-rT} N(d_2^n) \right), n \in \mathbb{N}^+    ,            \\
	\end{aligned}
	\label{eq:call}
\end{equation}
where $r$ is the risk-free rate.
and for put options:
\begin{equation}
	\begin{aligned}
		\tilde{V}_{E} \approx  \sum_{0}^{n} e^{-\lambda T} \frac{(\lambda T)^n}{n!} \left( K e^{-rT} N(-d_2^n) - S e^{n\mu_Y} N(-d_1^n) \right), \\
	\end{aligned}
	\label{eq:put}
\end{equation}
where
\begin{equation}
	d_1^n = \frac{\ln\left(\frac{S}{K}\right) + (r - \lambda k + n\mu_Y + 0.5\sigma_n^2)T}{\sigma_n \sqrt{T}},
	\label{eq:d1}
\end{equation}
and
\begin{equation}
	d_2^n = d_1^n - \sigma_n \sqrt{T},
	\label{eq:d2}
\end{equation}
with parameters defined as,
\begin{equation}
	\begin{cases}
		\sigma_n = \sqrt{\sigma^2 + \frac{n \sigma_Y^2}{T}},                  \\
		k = \mathbb{E}[e^{\ln Y} - 1] = e^{\mu_Y + \frac{\sigma_Y^2}{2}} - 1. \\
	\end{cases}
	\label{eq:parameters}
\end{equation}
where $r$ and $\sigma$ are trainable parameters in the neural network.

\textbf{Early Exercise Premium} \\
A fully connected feedforward neural network is used to estimate $\hat{\Pi}$, which contains two hidden layers with fixed features dimensions of 128.
For each hidden layer, after linear transformation, batch normalization and ReLU activation are applied for training stability;
Moreover, a dropout layer with rate of 0.3 is applied to prevent overfitting.
The output layer maps the hidden features to a single output, and further passing through a Softplus activation function to ensure a non-negative and smooth result.

Finally, the American Option Price is calculated by,
\begin{equation}
	\tilde{V} = a \cdot \tilde{V}_E + (1-a) \cdot \hat{\Pi}.
	\label{eq:american option price}
\end{equation}

\subsection{III.Analytical Merton Model}

Analytical formula for pricing European options was presented by Merton in \citep{merton1976option}, and the formulas are given in Equation \ref{eq:call} and \ref{eq:put}.
We present the change of measure here.
Given optimized parameters $\mu^*, \sigma^*, \lambda^*, \mu_Y^*, \sigma_Y^*$, the SDE of price process under real-world measure $\mathbb{P}$ can be expressed as,
\begin{equation}
	dS_t = \mu^* S_t dt + \sigma^* S_t dW_t^{\mathbb{P}} + S_t d(\sum_{i=1}^{N^{\mathbb{P}}_t}(Y_i-1))
	\label{eq:sde merton}
\end{equation}
Following GBM favor, we adjust the drift term $\mu^*$ from measure $\mathbb{P}$ to $\mu^{\dagger}$ under risk-neutral measure $\mathbb{Q}$ given risk-free rate $r$, defined as,
\begin{equation}
	\begin{aligned}
		\mu^{\dagger} = r - \lambda^* (e^{\mu_{Y^*} + \frac{\sigma^{2*}_{Y}}{2}}-1 ) \\
		= r - \lambda^* k^*
	\end{aligned}
	\label{eq:risk-neutral-drift}
\end{equation}
Then the SDE under risk nuetral measure can be expressed as,
\begin{equation}
	dS_t = \mu^{\dagger} S_t dt + \sigma^* S_t dW_t^{\mathbb{Q}} + S_t d(\sum_{i=1}^{N_t^{\mathbb{Q}}}(Y_i-1))
	\label{eq:sde merton Q}
\end{equation}
where $dW_t^{\mathbb{Q}}$ indicates standard GBM under $\mathbb{Q}$ measure,  and $N_t^{\mathbb{Q}}$ indicates the jump process with same jump intensity under $\mathbb{Q}$ measure.

\subsection{IV.Loss coefficients Optimization Module}

To balance the training efficiency and optimization effect, an loss coefficients optimization framework is degined in Algorithm \ref{Algorithm:loss coefficients}.
The \textit{Optuna} API proposed by \cite{akiba2019optuna} is employed, which utilizes Bayesian optimization with the \textit{Tree-structured Parzen Estimator}.

\begin{algorithm}[H]
	\SetAlgoLined
	\KwIn{Train dataloader $(X_{\text{train}}, y_{\text{train}})$, Validation dataloader $(X_{\text{val}}, y_{\text{val}})$}
	\KwOut{Optimal hyperparameters $\alpha^*$, $\beta^*$}
	Initialize $\mathcal{D}_0 = \{(\alpha_i, \beta_i, L(\alpha_i, \beta_i)) \mid i = 1, \ldots, n\}$ with $n$ initial points\\
	\For{iteration $k = 1$ to $K$}{
		Fit a probabilistic surrogate model $p(L \mid \alpha, \beta, \mathcal{D}_{k-1})$ to the observed data points\\
		Define an acquisition function $a(\alpha, \beta \mid \mathcal{D}_{k-1})$ to guide the search for the next hyperparameters to evaluate\\
		$(\alpha_k, \beta_k) = \arg \max_{(\alpha, \beta)} a(\alpha, \beta \mid \mathcal{D}_{k-1})$\\
		Evaluate the objective function $L(\alpha_k, \beta_k)$:
		\begin{enumerate}
			\item Initialize model, optimizer, and scheduler with $\alpha_k$ and $\beta_k$
			\item \For{epoch $e = 1$ to $E$}{
				      \begin{enumerate}
					      \item \For{each batch $(X_{\text{batch}}, y_{\text{batch}})$ in train dataloader}{
						            \begin{enumerate}
							            \item Perform forward pass to obtain predictions $\hat{y}_{\text{batch}}$
							            \item Compute total loss based on Equation (\ref{eq:loss})
							            \item Perform backward pass and update model parameters
						            \end{enumerate}
					            }
					      \item Update learning rate scheduler
				      \end{enumerate}
			      }
			\item Compute validation loss $L_{\text{val}}$ on validation dataloader
		\end{enumerate}
		Update the dataset $\mathcal{D}_k = \mathcal{D}_{k-1} \cup \{(\alpha_k, \beta_k, L_{\text{val}}(\alpha_k, \beta_k))\}$\\
	}
	\caption{Loss coefficients Optimization}
	\label{Algorithm:loss coefficients}
\end{algorithm}

\subsection{V. Numerical PIDE Solver Module}

This module is designed for data augmentation based on numerical PIDE Solver, the generated synthetic datasets are used for model pre-training.

\textbf{Mathematical Model}\\
The mathematical model of Merton jump diffusion process is governed by a Partial Integro-Differential Equation (PIDE):
\begin{equation}
	\begin{aligned}
		\frac{\partial V(t,x)}{\partial t} + \left(r - \frac{\sigma^2}{2} - \lambda \left( e^{\mu_Y + \frac{\sigma_Y^2}{2}} - 1 \right) \right) \frac{\partial V(t,x)}{\partial x} + \\
		\frac{\sigma^2}{2} \frac{\partial^2 V(t,x)}{\partial x^2} +  \int_{-\infty}^{\infty} \left[ V(t,x+y) - V(t,x) \right] \nu(dy) - rV(t,x) = 0
	\end{aligned}
	\label{eq:PIDE}
\end{equation}
where $x = \text{log}(S)$, and $\nu(dy)$ is the Lévy measure, which is defined using a scaled Normal distribution as:
\begin{equation}
	\nu(dy) = \frac{\lambda}{\sigma_Y \sqrt{2\pi}} e^{- \frac{(y-\mu_Y)^2}{2\sigma_Y^2}} dy.
	\label{eq:levy_measure}
\end{equation}
Moreover, the boundary conditions for American call and put options are defined as:
\begin{equation}
	\begin{cases}
		V(x_{\min}, t) = 0 \\
		V(x_{\max}, t) = S_{\max} - K e^{-r(T-t)}
	\end{cases}
	\label{eq:Merton_BC_C}
\end{equation}
and
\begin{equation}
	\begin{cases}
		V(x_{\min}, t) = K e^{-r(T-t)} - S_{\min} \\
		V(x_{\max}, t) = 0
	\end{cases}
	\label{eq:Merton_BC_P}
\end{equation}
respectively.
Additionally, the initial condition at maturity $t=T$ is defined as:
\begin{equation}
	\begin{cases}
		V_C(x, T) = \max(S - K, 0) \\
		V_P(x, T) = \max(K - S, 0)
	\end{cases}
	\label{eq:Merton_Initial}
\end{equation}

\textbf{Discretization}\\

To avoid the inversion of a dense jump matrix, an Implicit-Explicit (IMEX) scheme is employed where the differential part of Equation (\ref{eq:PIDE}) is discretized using an implicit scheme while the integral part is discretized using an explicit scheme.
To accurately cover the domain of the Lévy measure, the boundary regions of this IMEX scheme need to be extended, resulting in extra grid nodes on the boundaries.
Therefore, in the computational domain $[0,T] \times [L-E,U+E]$, where $L$, $U$, and $E$ are the lower, upper, and extended lines, the boundary nodes are located in the regions of $[0,T] \times [L-E,L]$ and $[0,T] \times [U,U+E]$, respectively.

A uniform grid is generated by spatial and temporal discretization as following:
\begin{equation}
	\begin{cases}
		x_i & = x_{\min} + i \Delta x \quad \text{for} \quad i = 0, 1, \ldots, N_x; \Delta x = \frac{x_{\max} - x_{\min}}{N_x}, N_x \in \mathbb{N}^+ \\
		t_j & = j \Delta t \quad \text{for} \quad j = 0, 1, \ldots, N_y; \Delta t = \frac{T}{N_y}, N_y \in \mathbb{N}^+                              \\
	\end{cases}
	\label{eq:Numerical_discretization}
\end{equation}

Moreover, Equation (\ref{eq:PIDE}) can be discretized and solved using the IMEX scheme.
Specifically, the discretized form of equations, the coefficients, and the formation of the tridiagonal matrix in shown Appendix B.
In addition, the \textit{Scipy} API is employed to solve the algebraic equations, and Fast Fourier Transform is used to compute convolution integral for jumps term update.

\subsection{VI. Training Module}

Two schemes are included in the training module, namely PINN-Merton and PINN-Merton-Transfer. 
The PINN-Merton is trained using the market datasets with scarce data, while the PINN-Merton-Transfer is trained using both the market datasets and the synthetic datasets generated by the numerical PIDE solver. 
Specifically, in the PINN-Merton-Transfer scheme, it first trains a PINN-Merton model using only the synthetic data, then freezes the parameters of the two hidden layers in the trained model and retrains the model using the market dataset.

For both schemes, mixed precision, early stopping, and a learning rate scheduler are used to enhance training performance. 
Specifically, the Super-Convergence techniques proposed by \cite{smith2019super} are used to update the learning rate in each epoch. 
In addition, to ensure the physical effectiveness of the framework, for both training schemes, hard constraints based on the empirical bounds of the physics parameters are introduced to clamp the parameters of the neural network-parametrized European option pricing module within a physically effective range.

\section{Experiments and Analysis}
\subsection{Dataset and Baselines}

The historical SPY close prices from January 1, 2010, to January 1, 2024 are obtained from \textit{Yahoo Finance} API for parameters optimization.
The data of SPY option chains from Q1 2020 to Q4 2022 are provided by Kyle Graupe on \textit{Kaggle}\footnote{https://www.kaggle.com/datasets/kylegraupe/spy-daily-eod-options-quotes-2020-2022}.
The raw data of call and put options are 3,589,079 each. The mean of the bid and ask prices are defined as target following \citet{dhiman2023physics}.
We employ the random sampling method provided by \textit{Pandas} to construct four scarce datasets (see Table \ref{tab:datasets}), where the random seed is fixed at 42.
Compared to previous studies using 205,123 samples \citep{jang2021deepoption}, our datasets are significantly smaller.

\begin{table}
	\begin{threeparttable}
		\caption{Description of Experiment Datasets}
		\setlength{\tabcolsep}{3pt}  % Adjust column spacing to optimize layout
		\begin{tabular}{l c c c c c c c}
			\toprule
			% Row headers
			\textbf{Cases}     & \textbf{Type} & \textbf{Sample Ratio} & \textbf{Train Samples} & \textbf{Validation Samples} & \textbf{Test Samples} \\
			\midrule
			Case 1             & Call          & 0.1\%                 & 2512                   & 538                         & 539                   \\
			Case 2             & Put           & 0.1\%                 & 2512                   & 538                         & 539                   \\
			Case 3             & Call          & 0.05\%                & 1256                   & 269                         & 270                   \\
			Case 4             & Put           & 0.05\%                & 1256                   & 269                         & 270                   \\
			Case 6*  & Put           & 100\%/0.1\%           & 70,000/2512            & 15,000/538                  & 15,000/539            \\
			Case 7*  & Call          & 100\%/0.1\%           & 70,000/2512            & 15,000/538                  & 15,000/539            \\
			\bottomrule
		\end{tabular}
		\label{tab:datasets}	
		\footnotesize{$*$ Transfer learning using synthetic and real datasets}
	\end{threeparttable}
\end{table}

\begin{figure}[H]
	\centering
	\includegraphics[trim=0cm 0cm 0cm 0cm,clip=true,width=15cm]{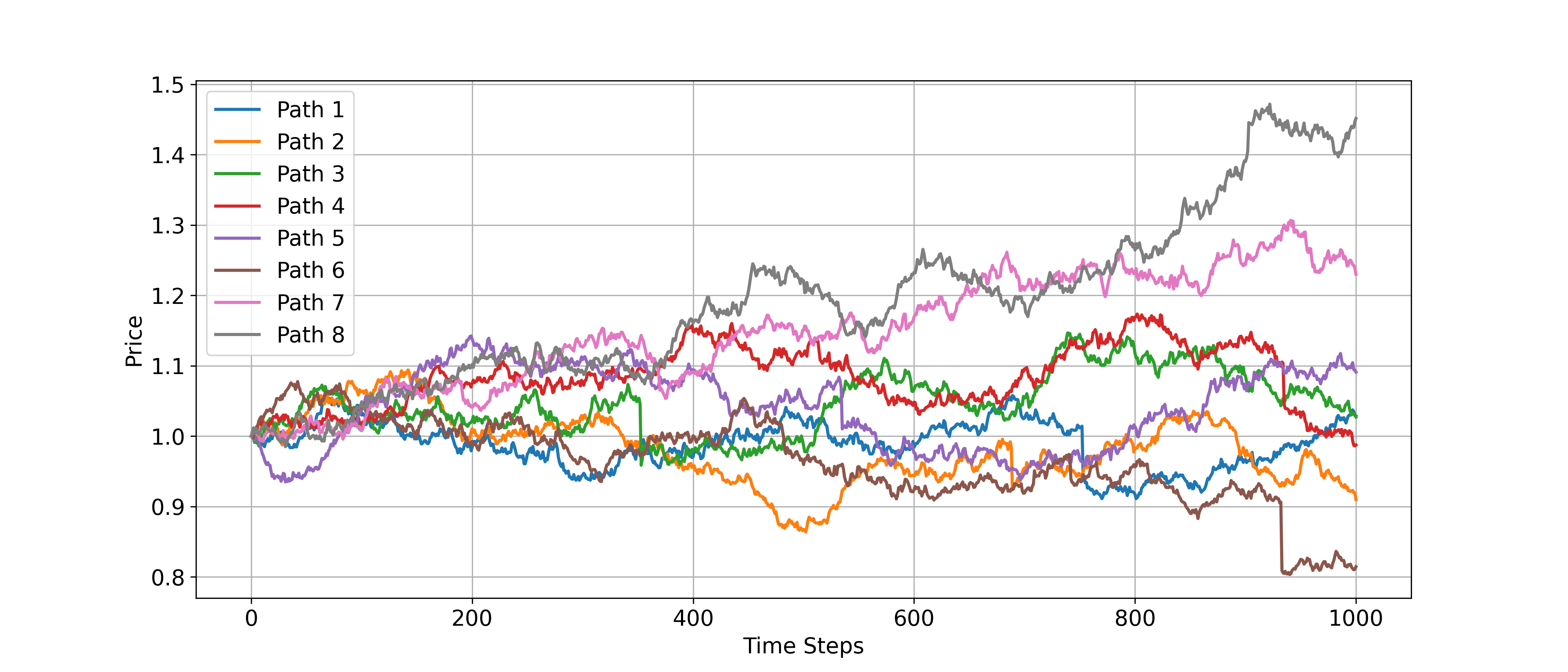}
	\vspace*{-3mm}
	\caption{Eight stochastic price paths generated by the Merton jump diffusion model with optimized parameters}
	\label{Fig:paths}
\end{figure}

Applying Algorithm \ref{Algorithm:loss coefficients} on the SPY prices data results in optimized parameters of $\mu = 0.179$, $\sigma = 0.143$, $\lambda = 2.0$, $\mu_Y = -0.012$, and $\sigma_Y = 0.042$.
Figure \ref{Fig:paths} shows eight generated paths using these parameters, and Figure \ref{Fig:paths} illustrates the comparison of the log returns between generated paths and real SPY data.
It can be visually confirmed that the generated log returns can accurately recover the true distributions, implying the Merton model can capture leptokurtosis distributions of log return series in real SPY data.                              

\begin{figure}[H]
	\centering
	\includegraphics[trim=0cm 0cm 0cm 0cm,clip=true,width=15cm]{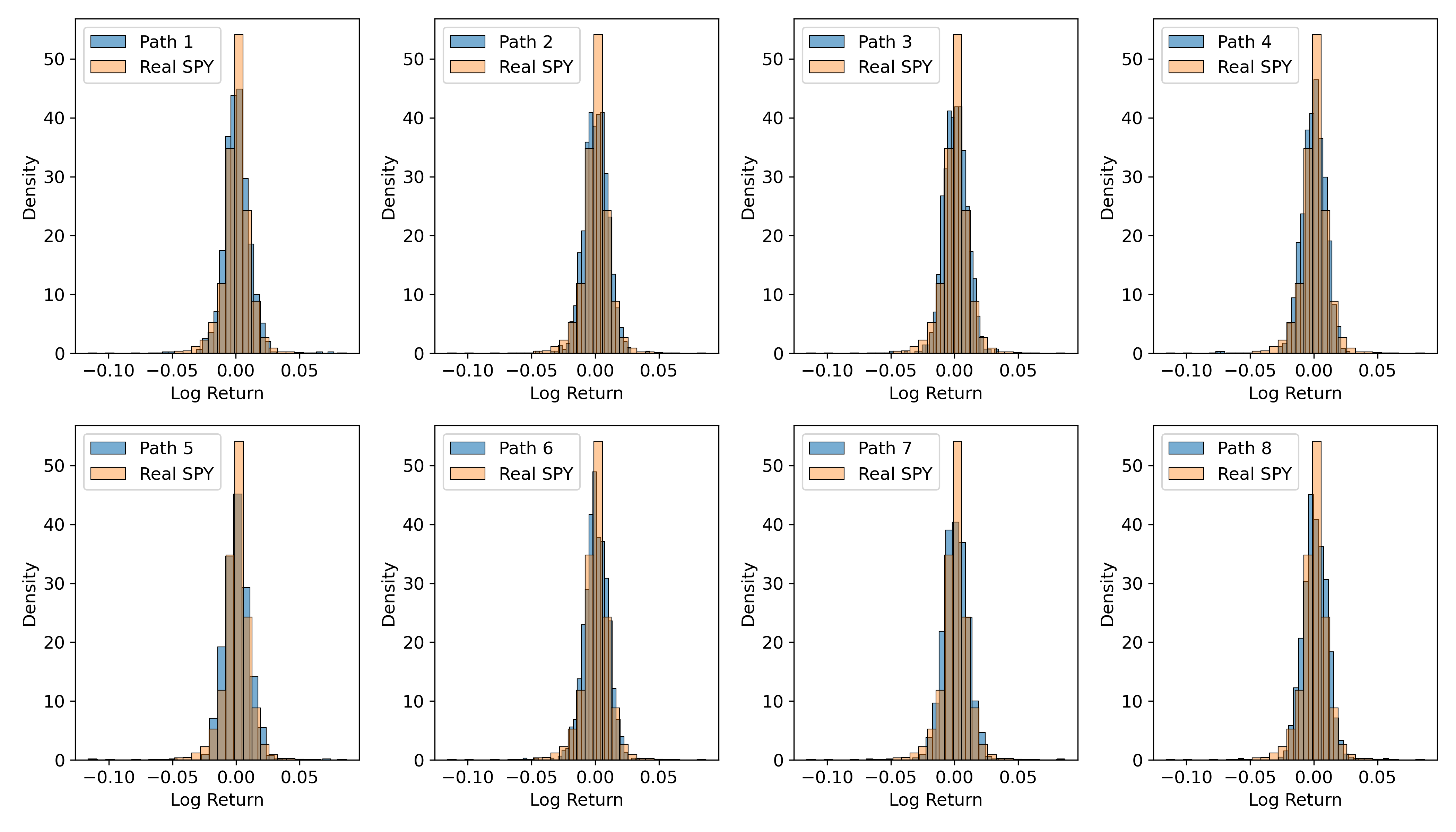}
	\vspace*{-3mm}
	\caption{Comparison of the Log return distributions of real SPY close price and the eight stochastic paths generated by the Merton jump diffusion model with optimized parameters}
	\label{Fig:paths distribution}
\end{figure}

To create synthetic datasets, we employ the change of measure defined in Equation (\ref{eq:risk-neutral-drift}) to obtain drift under measure $\mathbb{Q}$ and assign parameters to both analytical and numerical models.
Then the variable sets of $S$, $K$, $T$, and $r$ are generated following uniform distributed, then the numerical model is used to generate the corresponding targets.
Finally, a large synthetic dataset containing 100,000 call and 100,000 put options is generated.

Three baselines are considered for comparison: PINN-BS, Neural Networks (NN), and XGBoost.
The PINN-BS model is proposed in \citep{dhiman2023physics}, while the NN and XGBoost are investigated in \cite{IVASCU2021113799}.
In the experiments by \cite{IVASCU2021113799} with sufficient data, XGBoost shows significant performance compared to other baselines.
Additionally, although we do not compare with studies using more features as input, our framework is extendable to high-dimensional feature conditions.

\subsection{Framework Performance}

\begin{figure}[H]
	\centering
	\includegraphics[trim=0cm 0cm 0cm 0cm,clip=true,width=15cm]{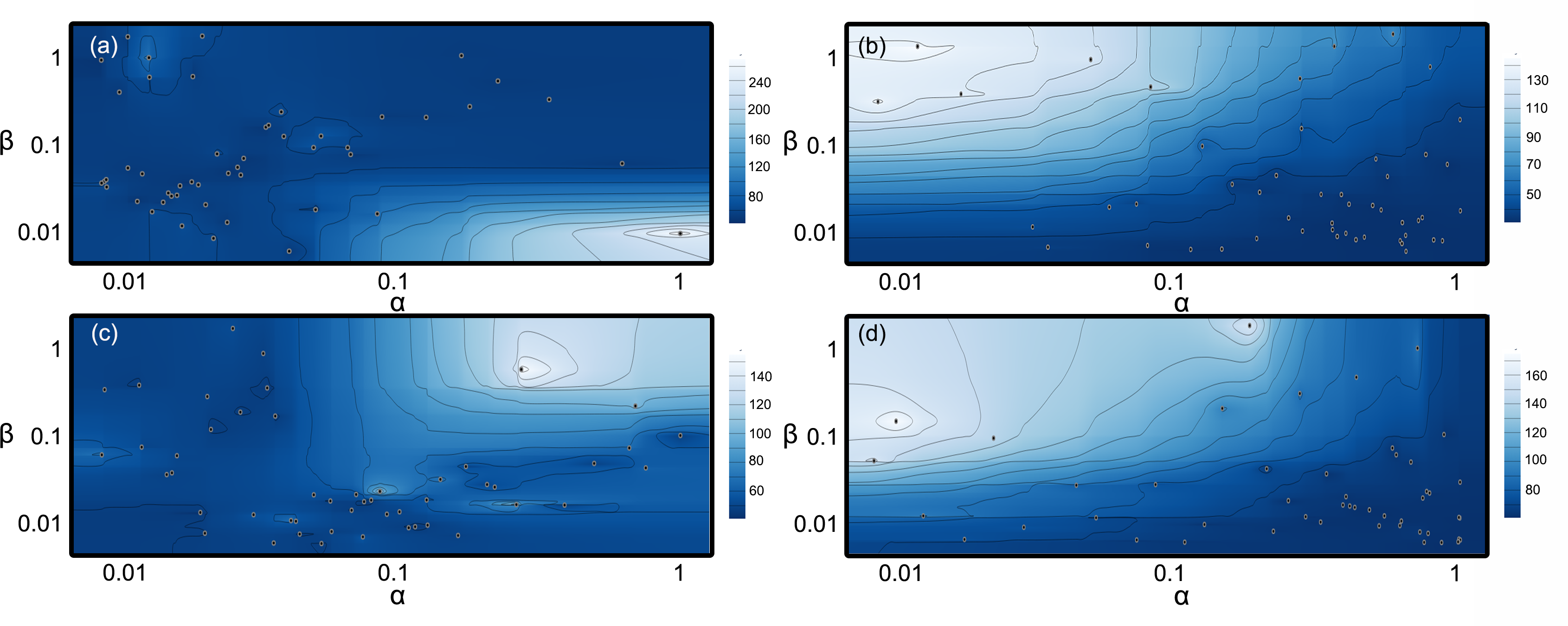}
	\vspace*{-3mm}
	\caption{Contour plot of validation losses under different physics loss $\alpha$ and data loss $\beta$ for (a) PINN-Merton for call options pricing; (b) PINN-Merton for put options pricing; (c) PINN-BS for call options pricing; (d) PINN-BS for put options pricing. The number of trials for Bayesian optimization and number of epochs in each trial are set to 50 and 50, respectively. }
	\label{Fig:contour}
\end{figure}

It is found that the training efficiency and accuracy of PINN-Merton are significantly influenced by the coefficients $\alpha^*$ and $\beta^*$ optimized through Algorithm \ref{Algorithm:loss coefficients}.
We set number of trials to 50, the number of epochs per trial to 100, and the bounds of $\alpha$ and $\beta$ as [0.01, 1] each.
The optimized validation losses of 50 trials are presented in Figure \ref{Fig:contour} with dark blue areas indicate low losses.

Overall, consistent distributions are observed in the PINN-Merton model in Figure \ref{Fig:contour} (a) and (b).
Specifically, when applying PINN-Merton for call options pricing, $\alpha$ and $\beta$ pairs are distributed along the diagonal, indicating that physical loss and data loss have similar weights.
When utilizing PINN-Merton to price put options, the optimal weight of physical loss $\alpha$ should be higher than that of data loss $\beta$.
However, the loss pairs distribution in PINN-BS model in Figure \ref{Fig:contour} (c) is ambiguous.
Finally, similar patterns are observed in Figure \ref{Fig:contour} (b) and (d), implying that when pricing American put options of SPY option chains, the weight of the physics loss should be set higher than that for the data loss for both PINN-Merton and PINN-BS models.

\begin{figure}[H]
	\centering
	\includegraphics[trim=0cm 0cm 0cm 0cm,clip=true,width=15cm]{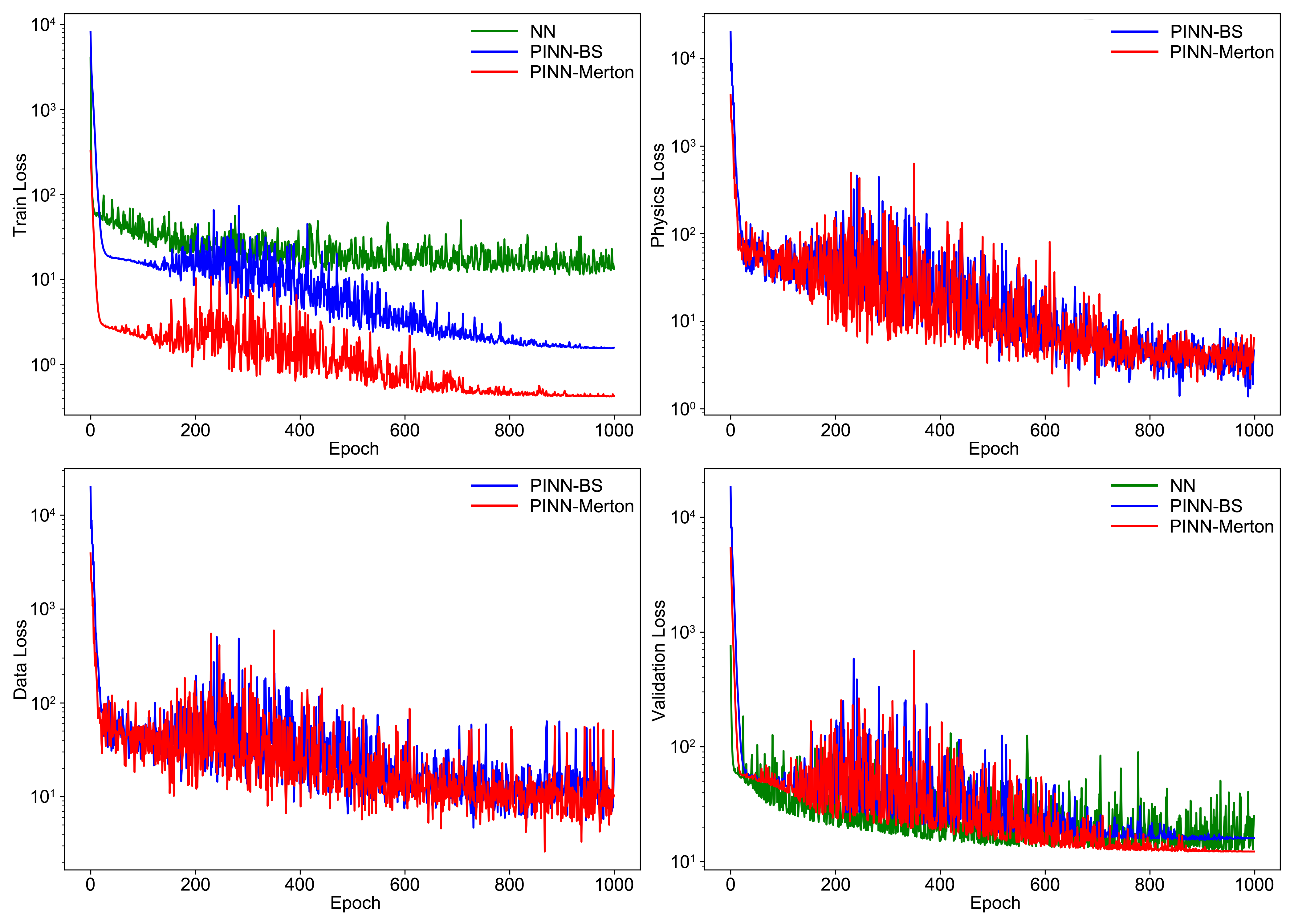}
	\vspace*{-3mm}
	\caption{Training loss, validation loss, physics loss, and data loss for PINN-Merton, PINN-BS, and NN models during 1000 training epochs on call options datasets with a sample ratio of 0.1\%.}
	\label{Fig:lossesC}
\end{figure}

\begin{figure}[H]
	\centering
	\includegraphics[trim=0cm 0cm 0cm 0cm,clip=true,width=15cm]{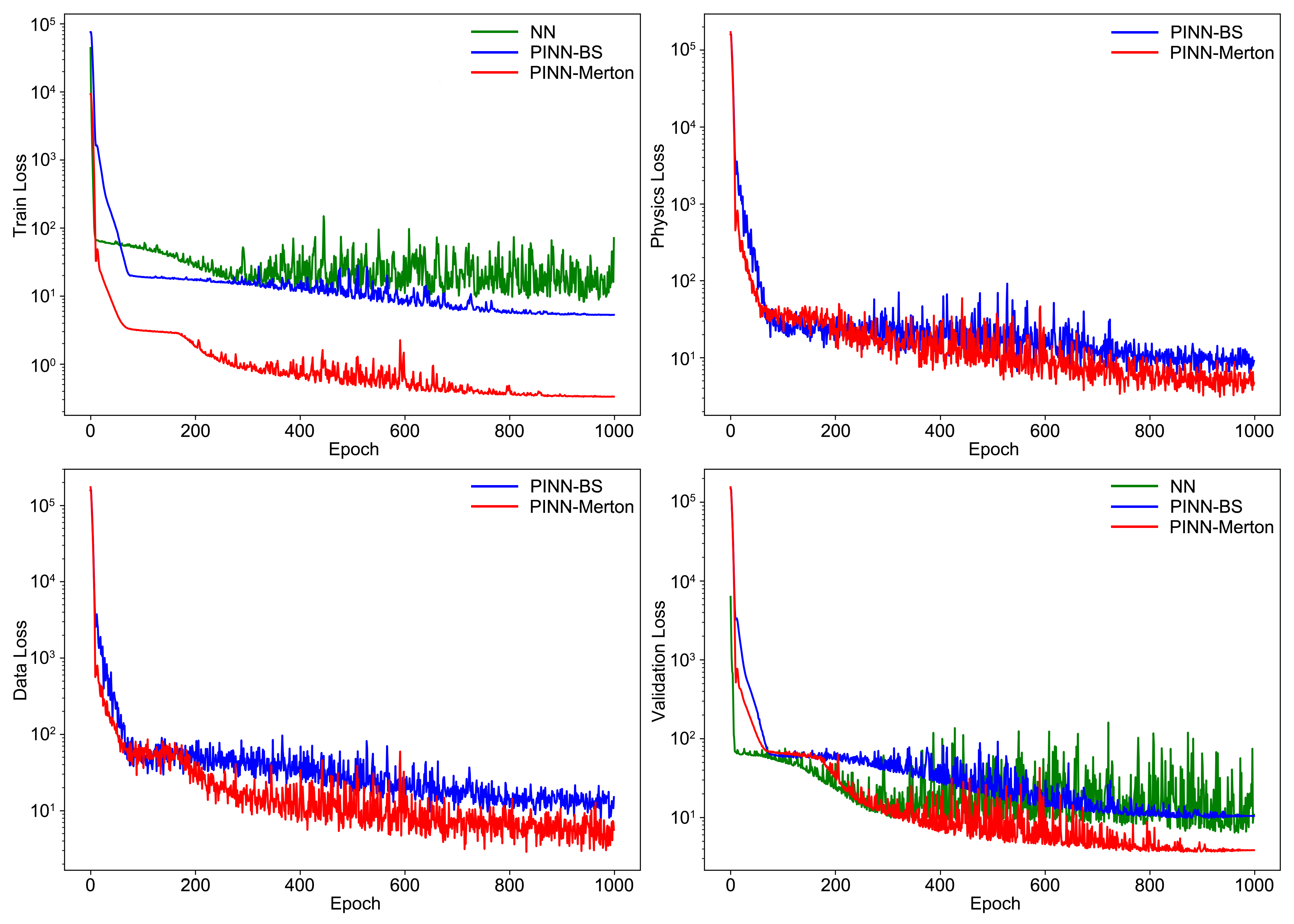}
	\vspace*{-3mm}
	\caption{Training loss, validation loss, physics loss, and data loss for PINN-Merton, PINN-BS, and NN models during 1000 training epochs on put options datasets with a sample ratio of 0.05\%.}
	\label{Fig:lossesP}
\end{figure}

Figures \ref{Fig:lossesC} and \ref{Fig:lossesP} illustrate the training loss, physics loss, data loss, and validation loss for PINN-Merton, PINN-BS, and NN models for call and put options pricing tasks, respectively.
Generally, PINN-Merton outperforms PINN-BS and NN models in terms of training and validation losses. Moreover, after 900 epochs, the training and validation losses in PINN-Merton and PINN-BS tend to converge, which cannot be observed in the NN model.
Since the neural network architecture of the three models is similar, it demonstrates the convergences of the PINN models are better than the neural network model only.
Furthermore, in Figure \ref{Fig:lossesC}, the physical loss and data loss of PINN-Merton and PINN-BS almost overlap, while in Figure \ref{Fig:lossesP}, the physical loss and data loss of PINN-Merton are significantly lower than those of PINN-BS, indicating that PINN-Merton is more effective in pricing American put options.

\begin{figure}[H]
	\centering
	\includegraphics[trim=0cm 0cm 0cm 0cm,clip=true,width=15cm]{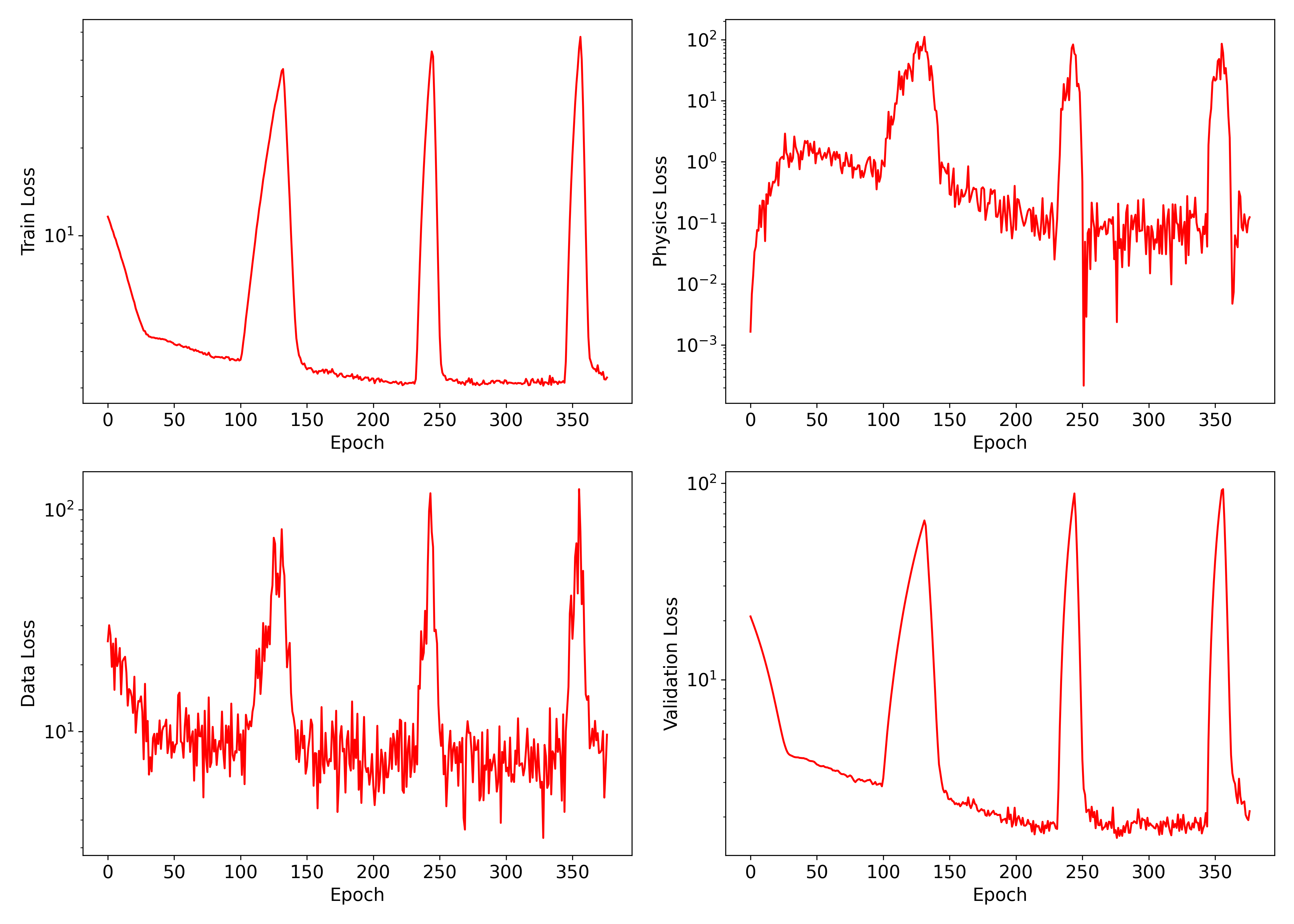}
	\vspace*{-3mm}
	\caption{Training loss, validation loss, physics loss, and data loss for PINN-Merton-Transfer model, the pre-train model is based on the synthetic dataset.}
	\label{Fig:losses-transfer}
\end{figure}

Figure \ref{Fig:losses-transfer} shows very interesting simultaneous periodic patterns in the training loss, physics loss, data loss, and validation loss for PINN-Merton-Transfer. 
This phenomenon results from the 1cycle learning rate policy, which anneals the learning rate from an initial value to a maximum value and then reduces it to a minimum value much lower than the initial rate. 
It can be seen that the loss troughs gradually decrease during the training process. Therefore, by adopting the early stopping technique, the best model can be captured during the training process.

Moreover, note that benefiting from the use of transfer learning, the initial losses in this figure are very low compared with Figures \ref{Fig:lossesC} and \ref{Fig:lossesP},
indicating strong convergence of PINN-Merton-Transfer. 
Additionally, it can be seen that the magnitude of the physics loss is much less than that in the PINN-Merton schemes, which implies that PINN-Merton-Transfer is more physically effective compared to PINN-Merton.

The performance of all models on test datasets is summarized in Table \ref{tab: fully datasets}.
Overall, in Cases 1-4, where only real datasets with scarce data are used, PINN-Merton achieves the best performance on all metrics under different conditions, indicating that the Merton jump diffusion model is better than the BS model and pure data-driven models in capturing real price movement.

Moreover, comparing PINN-Merton-Transfer in Case 5 with PINN-Merton in Case 1 for call options pricing, as well as PINN-Merton-Transfer in Case 6 with PINN-Merton in Case 2 for put options pricing, 
it can be found that the PINN-Merton-Transfer significantly outperforms PINN-Merton after integrating transfer learning using synthetic datasets,
implying better accuracy and strong generalization of PINN-Merton-Transfer on scarce data conditions.

We also compare PINN-Merton performance with DeepOption \citep{jang2021deepoption} on S\&P call options datasets, and the best performance of DeepOption is reported with an MAE of 7.0 and an RMSE of 11.0, which is competitive compared to PINN-Merton.
However, it should be noted that the training dataset used in our work is much smaller than the dataset used for DeepOption training.

\begin{longtable}{lcccccc}
\caption{Performance of Various Models for American Option Pricing on Test Datasets}\\
\toprule
\textbf{Case-Model} & \textbf{MAE} & \textbf{MSE} & \textbf{RMSE} & \textbf{R2 Score} & \textbf{EV} & \textbf{Max Error} \\
\midrule
\endfirsthead
\multicolumn{7}{c}%
{{\bfseries \tablename\ \thetable{} -- continued from previous page}} \\
\toprule
\textbf{Case-Model} & \textbf{MAE} & \textbf{MSE} & \textbf{RMSE} & \textbf{R2 Score} & \textbf{EV} & \textbf{Max Error} \\
\midrule
\endhead
\midrule \multicolumn{7}{r}{{Continued on next page}} \\
\endfoot
\bottomrule
\endlastfoot
Case1-PINN-Merton & 1.9775 & 10.5236 & 3.2440 & 0.9977 & 0.9977 & 14.9531 \\
Case1-PINN-BS & 2.4318 & 15.0135 & 3.8747 & 0.9967 & 0.9969 & 26.9812 \\
Case1-NN & 3.6940 & 22.0236 & 4.6929 & 0.9951 & 0.9975 & 17.7096 \\
Case1-XGBoost & 4.7336 & 51.9335 & 7.2065 & 0.9873 & 0.9873 & 52.5186 \\
\hline 
Case2-PINN-Merton & 2.1509 & 9.7425 & 3.1213 & 0.9957 & 0.9961 & 12.3597 \\
Case2-BS & 3.0797 & 16.4402 & 4.0547 & 0.9928 & 0.9928 & 13.2692 \\
Case2-NN & 2.6931 & 14.2106 & 3.7697 & 0.9938 & 0.9946 & 13.1531 \\
Case2-XGBoost & 3.9559 & 59.9099 & 7.7401 & 0.9759 & 0.9760 & 80.6794 \\
\hline 
Case3-PINN-Merton& 1.9826 & 9.8792 & 3.1431 & 0.9974 & 0.9974 & 13.8444 \\
Case3-BS & 2.3142 & 13.3277 & 3.6507 & 0.9965 & 0.9966 & 16.3452 \\
Case3-NN  & 2.8796 & 12.8746 & 3.5881 & 0.9971 & 0.9979 & 14.0032 \\
Case3-XGBoost  & 6.3793 & 76.1452 & 8.7261 & 0.9780 & 0.9782 & 37.7978 \\
\hline 
Case4-PINN-Merton& 1.4462 & 6.0969 & 2.4692 & 0.9971 & 0.9971 & 13.7209 \\
Case4-BS & 2.4918 & 14.7215 & 3.8369 & 0.9930 & 0.9931 & 18.5296 \\
Case4-NN& 2.6743 & 13.3849 & 3.6585 & 0.9936 & 0.9952 & 14.9076 \\
Case4-XGBoost & 4.5213 & 63.0418 & 7.9399 & 0.9664 & 0.9669 & 75.5599 \\
\hline
Case5-PINN-Merton & 1.6597 & 7.4135 & 2.7228 & 0.9984 & 0.9984 & 12.5505 \\
Case6-PINN-Merton & 1.4338 & 4.9329 & 2.2210 & 0.9978 & 0.9981 & 11.2906 \\
\label{tab: fully datasets}
\end{longtable}

The performance of all models for pricing deep out-of-the-money American options ($V<10$) is summarized in Table \ref{tab: otm datasets}.
In practice, pricing these out-of-the-money options is challenging because they have a low probability of being exercised due to their strike price being far from the current underlying asset price. 
Moreover, these deep out-of-the-money options often lack liquidity. Furthermore, sudden changes in market conditions disproportionately affect the prices of these options.

Nevertheless, it can be seen that PINN-Merton and PINN-Merton-Transfer achieve reasonable performance for pricing both call and put options with sample ratios of 0.1\%, with PINN-Merton being slightly inferior to PINN-Merton-Transfer. 
However, when sample ratios decreased to 0.05\%, the performance of PINN-Merton significantly decreased for pricing call options. 
Additionally, the other two models, NN and XGBoost, show poor performance in pricing these deep out-of-the-money options.

\begin{longtable}{lcccccc}
\caption{Performance of Various Models for Deep Out-of-the-Money American Option Pricing on Test Datasets}\\
\toprule
\textbf{Case-Model} & \textbf{MAE} & \textbf{MSE} & \textbf{RMSE} & \textbf{R2 Score} & \textbf{EV} & \textbf{Max Error} \\
\midrule
\endfirsthead

\multicolumn{7}{c}%
{{\bfseries \tablename\ \thetable{} -- continued from previous page}} \\
\toprule
\textbf{Case-Model} & \textbf{MAE} & \textbf{MSE} & \textbf{RMSE} & \textbf{R2 Score} & \textbf{EV} & \textbf{Max Error} \\
\midrule
\endhead

\midrule \multicolumn{7}{r}{{Continued on next page}} \\
\endfoot

\bottomrule
\endlastfoot
Case1-PINN-Merton& 1.2819 & 3.9982 & 1.9995 & 0.3554 & 0.5012 & 12.4592 \\
Case1-PINN-BS & 2.1498 & 10.5422 & 3.2469 & -0.6995 & -0.0587 & 20.2758 \\
Case1-NN & 2.9598 & 11.0433 & 3.3231 & -0.7803 & 0.3997 & 8.3204 \\
Case1-XGBoost& 3.4915 & 31.7882 & 5.6381 & -3.4426 & -2.5273 & 39.9472 \\
\hline
Case2-PINN-Merton & 1.1716 & 2.8902 & 1.7000 & 0.6446 & 0.6750 & 5.7967 \\
Case2-PINN-BS & 2.7339 & 12.2516 & 3.5002 & -0.5066 & 0.2889 & 11.8856 \\
Case2-NN & 1.4845 & 4.1302 & 2.0323 & 0.4921 & 0.5181 & 6.9492 \\
Case2-XGBoost & 1.8460 & 10.0123 & 3.1642 & -0.2079 & -0.0153 & 24.9450 \\
\hline
Case3-PINN-Merton & 1.4364 & 4.4515 & 2.1099 & -0.0116 & 0.3148 & 7.0504 \\
Case3-PINN-BS & 1.8870 & 8.2619 & 2.8743 & -0.8774 & -0.2325 & 10.1873 \\
Case3-NN & 2.5297 & 9.1873 & 3.0311 & -0.4811 & 0.4633 & 8.7701 \\
Case3-XGBoost & 4.4022 & 34.2158 & 5.8494 & -4.2650 & -1.9796 & 22.5527 \\
\hline
Case4-PINN-Merton & 0.8574 & 2.2501 & 1.5000 & 0.6719 & 0.6971 & 8.0345 \\
Case4-PINN-BS & 1.4024 & 5.4000 & 2.3238 & 0.2126 & 0.2348 & 8.9478 \\
Case4-NN& 1.6717 & 3.8242 & 1.9555 & 0.4424 & 0.5724 & 4.9066 \\
Case4-XGBoost& 2.5177 & 19.5495 & 4.4215 & -1.2039 & -0.8460 & 32.7657 \\
\hline
Case5-PINN-Merton & 1.0759 & 2.8166 & 1.6783 & 0.5459 & 0.6486 & 10.4574 \\
Case6-PINN-Merton & 1.0148 & 2.5208 & 1.5877 & 0.6900 & 0.7318 & 5.0251 \\
\label{tab: otm datasets}
\end{longtable}

\begin{figure}[H]
	\centering
	\includegraphics[trim=0cm 0cm 0cm 0cm,clip=true,width=14cm]{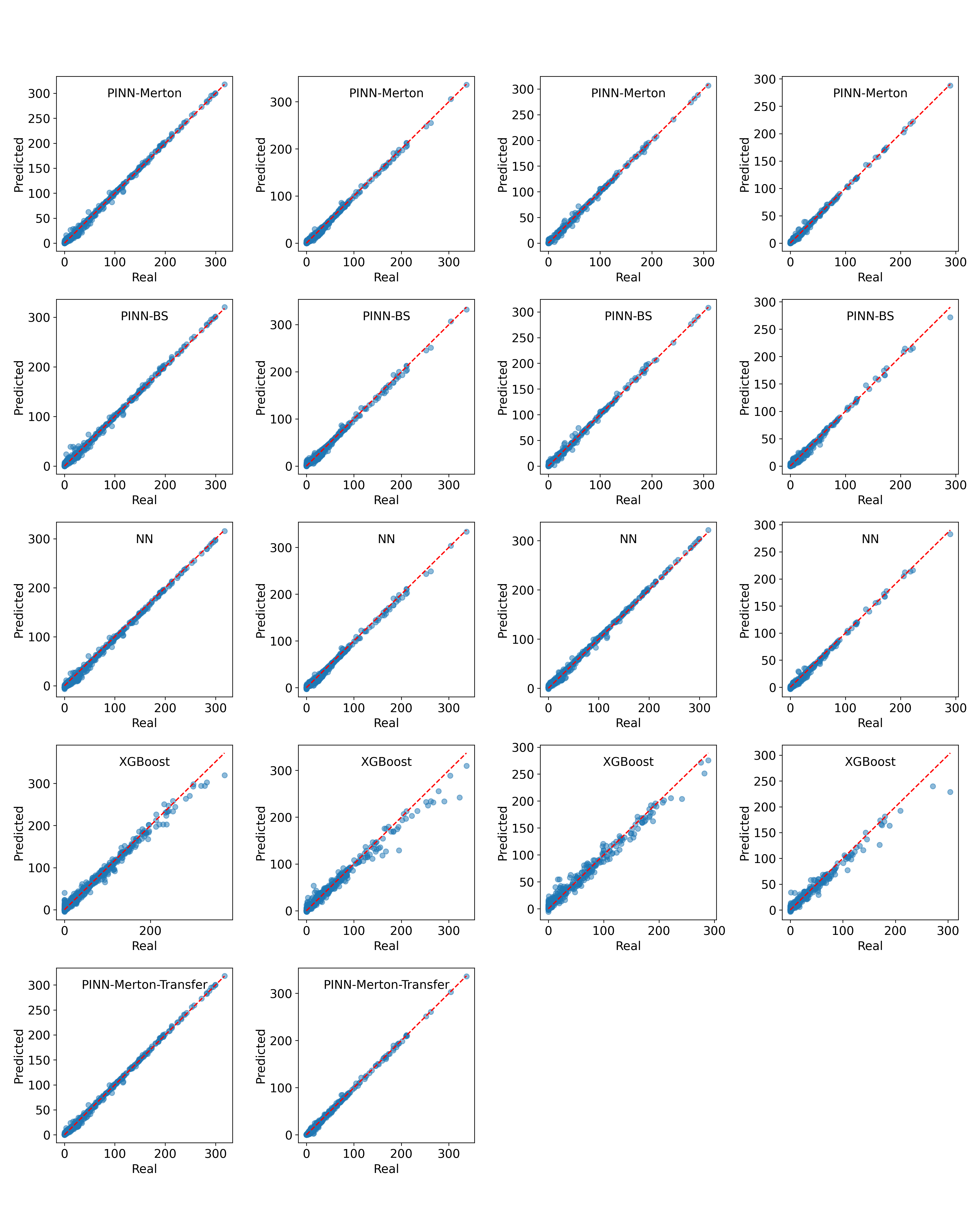}
	\vspace*{-10mm}
	\caption{Predicted vs Actual call options prices from PINN-Merton, PINN-Merton-Transfer, PINN-BS, NN, and XGBoost models on test datasets}
	\label{Fig:scatters}
\end{figure}

Figures \ref{Fig:scatters} further illustrate the predicted and actual option prices on real test datasets based on the five models. 
Generally, the PINN-Merton and PINN-Merton-Transfer models outperform the three baselines. Moreover, by integrating transfer learning techniques, PINN-Merton-Transfer further enhances the accuracy of PINN-Merton.
Additionally, since the PINN-Merton, PINN-BS, and NN models all implement similar neural network architectures, apart from the different parametric models used, it can be concluded that the significant performance of PINN-Merton is attributed to the effectiveness of integrating the jump diffusion process to model price processes, which better captures the leptokurtic nature of return series.
Furthermore, it is found that the point distribution from the XGBoost model deviates from the diagonal. When the sample ratio decreased to 0.05\%, the distribution further deviated and diffused, indicating the poor accuracy and generalization ability of XGBoost under scarce data conditions.
This finding is contrary to the experimental results of \citep{IVASCU2021113799}, indicating that pure data-driven models such as XBboost perform poorly in scarce data conditions.

\begin{figure}[H]
	\centering
	\includegraphics[trim=0cm 0cm 0cm 0cm,clip=true,width=14cm]{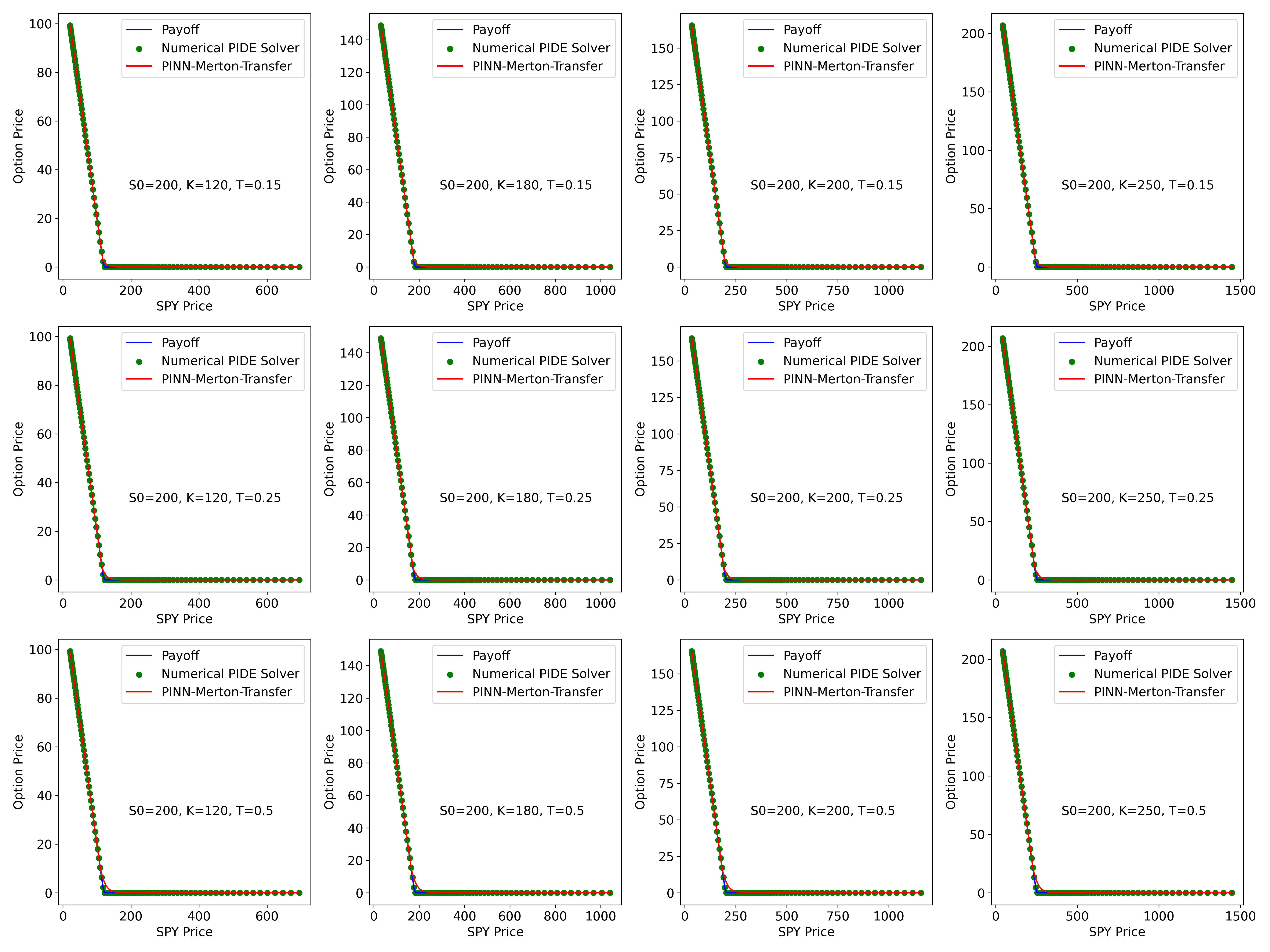}
	\vspace*{-3mm}
	\caption{PINN-Merton-Transfer and numerical model performance on the gird data}
	\label{Fig:num_calls}
\end{figure}

To further test the physical effectiveness and generalization of the trained PINN-Merton-Transfer model, a challenging task for option pricing on the grid was designed. This task is used only to evaluate theoretical model performance in previous studies.
In Figure \ref{Fig:num_calls}, the estimation of PINN-Merton-Transfer is represented by a red line, while the numerical model solution is indicated by green points. Generally, the estimations of the PINN-Merton-Transfer and the numerical solutions are very close under different monetary conditions. Since the considered exercise time is small, both estimated prices approach the real payoff line. 
Moreover, it can be seen that with the increase in t, the prices estimated by PINN-Merton-Transfer show a more significant deviation from the real payoff line near the strike prices region compared to numerical solutions, effectively reflecting the real prices distribution.
The results demonstrate that the PINN-Merton-Transfer, as a data-driven method, exhibits strong physical effectiveness and generalization in American option pricing tasks.

\section{Conclusion}

In this paper, we develop a comprehensive framework to address the inherent challenges in American option pricing under scarce data conditions.
The framework consists of six interrelated modules, integrating nonlinear optimization algorithms, analytical models, numerical models, and neural networks.
These modules collectively optimize parameter estimation, enhance physical effectiveness by introducing physical constraints, provide data augmentation, and increase model performance.
Moreover, transfer learning is incorporated into the training module to further improve pricing accuracy and physical effectiveness.
Additionally, a warm-up period for coefficients optimization is designed to balance data loss and physical loss during training, thereby enhancing the model's training efficiency and overall performance.

Six case studies are designed to examine the performance of the proposed framework.
Comparative results show that PINN-Merton outperforms traditional models and captures the leptokurtic nature of real return series.
By integrating transfer learning using synthetic datasets generated by numerical models, the PINN-Merton-Transfer further improves the pricing accuracy of the PINN-Merton model.
The experimental results on challenging grid data tasks demonstrate the physical effectiveness and generalization of PINN-Merton-Transfer.

Future work could explore more advanced stochastic volatility models to better capture real price processes, as well as extend the framework to other derivative pricing and risk management applications.

\appendix
\section{Merton Jump Diffusion Model for European Option Pricing}

Based on Equation (\ref{eq:Nt}), the change rate of the number of jumps $dN_t$ can be expressed as,
\begin{equation}
	dN_t =
	\begin{cases}
		1, & \text{with probability } \lambda dt   \\
		0, & \text{with probability } 1-\lambda dt \\
	\end{cases}
	\label{eq:DNt}
\end{equation}

By assuming independence of jump arrivals and jump size, the jump term can be formed as $(Y_t-1)dN_t$.
After removing the drift resulting from the jumps, a pure jump expression can be expressed as,

\begin{equation}
	\frac{dS_t}{S_t} = (Y_t-1)dN_t - \mathbb{E}[(Y_t-1)dN_t] = (Y_t-1)dN_t - \lambda k dt
\end{equation}
where $k = \mathbb{E}[(Y_t-1)]$. Moreover, combining GBM with jumps and considering multiple jumps conditions,
the price process can be further expressed as,
\begin{equation}
	\frac{dS_t}{S_t} = \mu dt + \sigma dW_t + \left( \prod_{j=1}^{dN_t} Y_j - 1 \right) - \lambda k dt
	\label{eq:jd1}
\end{equation}

To apply Itô's lemma, we reform Equation (\ref{eq:jd1}) as,
\begin{equation}
	dS_t = (\mu-\lambda k) S_t dt + \sigma S_t dW_t + S_t \left( \prod_{j=1}^{dN_t} Y_j - 1 \right)
	\label{eq:jd2}
\end{equation}

Applying Itô's lemma  \citep{ito1951stochastic} to the log function of price $S_t$ to transform the production form of jumps to summation form, resulting in,

\begin{equation}
	d \ln S_t = \frac{1}{S_t} dS_t - \frac{1}{2} \frac{1}{S_t^2} dS_t^2 + \sum_{j=1}^{dN_t} \ln Y_j
	\label{eq:jd3}
\end{equation}

Integrate above Equation and define 0 as the initial value of the Brownian motion and Poisson process, resulting in the following equation,

\begin{equation}
	\begin{aligned}
		\ln S_t - \ln S_0 = (\mu - \lambda k - \frac{1}{2} \sigma^2)t + \sigma (W_t - W_0) + \sum_{j=1}^{N_t - N_0} \ln Y_j \\
		= (\mu - \lambda k - \frac{1}{2} \sigma^2)t + \sigma W_t + \sum_{j=1}^{N_t} \ln Y_j
		\label{eq:jd5}
	\end{aligned}
\end{equation}

$\ln(Y_j)$ are normally distributed according to equation \ref{eq:yt}, thus the distribution of jump term in equation \ref{eq:jd5} can be expressed as $\sum_{j=1}^{N_t} \ln Y_j \sim \mathcal{N}(n \mu_Y, n \sigma_Y^2)$.
Moreover, the GBM process is assumed to follow $\sigma W_t \sim \mathcal{N}(0, \sigma^2 t)$. We can merge the GBM and jump term as and use a standard Normal distribution $Z$ to model the process as,

\begin{equation}
	\begin{aligned}
		\sigma W_t + \sum_{j=1}^{N_t} \ln Y_j \sim \mathcal{N}(n \mu_Y, (\frac{n \sigma_Y^2}{t} + \sigma^2) t) \\
		\sim n \mu_Y + \sqrt{\frac{n \sigma_Y^2}{t} + \sigma^2} \sqrt{t} Z                                     \\
		\sim n \mu_Y + \sqrt{\frac{n \sigma_Y^2}{t} + \sigma^2} W_t
	\end{aligned}
	\label{eq:merged}
\end{equation}

Next, define $\sigma_n = \sqrt{\frac{n \sigma_Y^2}{t} + \sigma^2}$ and substitute Equation (\ref{eq:merged}) into Equation (\ref{eq:jd5}),
\begin{equation}
	\begin{aligned}
		\ln S_t - \ln S_0 = ( n \mu_Y + \frac{n \sigma_Y^2}{2}) t + (\mu - \lambda k +\frac{\sigma_n^2}{2}) t + \sigma_n W_t \\
		\label{eq:jd6}
	\end{aligned}
\end{equation}

Exponentiate both sides and organize the initial price term, the update equation can be expressed as,

\begin{equation}
	\begin{aligned}
		S_t = S_0 e^{\left( n \mu_Y + \frac{n \sigma_Y^2}{2} \right)  + \left( \mu - \lambda k + \frac{\sigma_n^2}{2} \right) t + \sigma_n W_t} \\
		= S_0^{(n)} e^{\left( \mu- \lambda k - \frac{\sigma_n^2}{2} \right) t + \sigma_n W_t}                                                   \\
		\label{eq:jd7}
	\end{aligned}
\end{equation}

Recall in Equation (\ref{eq:risk-neutral-drift}) the drift term $\mu^{\dagger}$ under risk netural measure is defined.
Therefore, Equation (\ref{eq:jd7}) can be writen as $	S_t	= S_0^{(n)} e^{\left( \mu^{\dagger}- \frac{\sigma_n^2}{2} \right) t + \sigma_n W_t^{\mathbb{Q}} }$ under risk-nuetral measure $\mathbb{Q}$.
Finally, the European option pricing formula used in BS model can be introduced in pricing options with Merton jump diffusion process under $\mathbb{Q}$ measure.
Specifically, the 3 parameters generated in the above derivation can be defined as:
\begin{equation}
	\begin{cases}
		\sigma_n = \sqrt{\sigma^2 + \frac{n \sigma_Y^2}{T}}                  \\
		S_0^{n} = S_0 e^{ n \mu_Y + \frac{n \sigma_Y^2}{2}}                  \\
		k = \mathbb{E}[e^{\ln Y} - 1] = e^{\mu_Y + \frac{\sigma_Y^2}{2}} - 1 \\
	\end{cases}
	\label{eq:parameters2}
\end{equation}

Additionally, in real implementation, we replace conditional probabilities with expectations weighted by Poisson density and replace infinite series with truncated N-level series.
In this way, the European call options price under Merton jump diffusion process can be defined as,
\begin{equation}
	\begin{aligned}
		V(S_0^n,T) & = \sum_{n=0}^\infty V(S_0^n, T \mid N_T = n) P[N_T = n]                         \\
		           & = \sum_{n=0}^N V(S_0^n, T \mid N_T = n) \frac{(\lambda T)^n}{n!} e^{-\lambda T}
		\label{eq:pricing1}
	\end{aligned}
\end{equation}

\section{Numerical Merton PIDE Updating Function}
The updation function for American put options are given by,
\begin{equation}
	\begin{aligned}
		V_i^{j} = \max \Bigg( \max(K-S_i, 0), a_i V_{i-1}^{j+1} + b_i V_i^{j+1} + c_i V_{i+1}^{j+1} \\
		+ \Delta t \sum_{k=-\infty}^{\infty} \left[ V(x_i + k \Delta x, t_{j+1}) - V(x_i, t_{j+1}) \right] \nu(k \Delta x) \Bigg)
	\end{aligned}
	\label{eq:update_C}
\end{equation}
where $\nu(k \Delta x)$ denotes the discretized jump size distribution. For put options it can simply replace the payoff function of $ \max(S_i - K, 0)$ by   $ \max(K - S_i, 0)$.
Finally, the coefficients $a_i, b_i ,c_i  $ are given by,
\begin{equation}
	\begin{cases}
		a_i = \frac{\Delta t}{2} \left( \frac{\sigma^2}{\Delta x^2} - \frac{(r - \frac{\sigma^2}{2} - \lambda (e^{\mu_J + \frac{\sigma_J^2}{2}} - 1))}{\Delta x} \right) \\
		b_i = 1 + \Delta t \left( \frac{\sigma^2}{\Delta x^2} + r + \lambda \right)                                                                                      \\
		c_i = \frac{\Delta t}{2} \left( \frac{\sigma^2}{\Delta x^2} + \frac{(r - \frac{\sigma^2}{2} - \lambda (e^{\mu_J + \frac{\sigma_J^2}{2}} - 1))}{\Delta x} \right) \\
	\end{cases}
	\label{eq:num_coef}
\end{equation}

\bibliographystyle{elsarticle-num}
\bibliography{mybib}

\end{document}